\documentclass{article}
\usepackage[accepted]{icml2018}

\usepackage{verbatim}
\usepackage{microtype}
\usepackage{booktabs} % for professional tables
\usepackage{amsmath}
\usepackage{amsthm}
\usepackage{amsfonts}
\usepackage{amssymb}
\usepackage{comment}
\usepackage[bottom]{footmisc}
\usepackage[hyphens,spaces,obeyspaces]{url}
\usepackage[colorlinks=true, allcolors=blue]{hyperref}
\usepackage[inline]{enumitem}
\usepackage{subfig}
\usepackage{graphicx}
\usepackage{capt-of}
\usepackage{varwidth}
\usepackage[hyphens,spaces,obeyspaces]{url}

\usepackage{amssymb,amsthm,amsfonts,amscd,bbold}
\newlist{inlinelist}{enumerate*}{1}
\setlist*[inlinelist,1]{%
	label=(\roman*),
}

\newcommand{\A}{{A}}
\newcommand{\B}{{B}}
\newcommand{\Z}{{Z}}
\newcommand{\Za}{{Z}_a}
\newcommand{\Zb}{{Z}_b}
\newcommand{\F}{G_{\A\B}} % changed all to G (generator)
\newcommand{\G}{G_{\B\A}} 
\newcommand{\Fz}{G_{\A\B}} % note: removed ^z
\newcommand{\Gz}{G_{\B\A}}
\newcommand{\pdata}{p_d}
\newcommand{\pz}{p}
\newcommand{\Ea}{E_{A}}
\newcommand{\Eb}{E_{B}}

\newcommand{\loss}{\mathcal{L}}
\DeclareMathOperator*{\E}{\mathbb{E}}
 
\DeclareMathOperator*{\argmin}{arg\,min}

\icmltitlerunning{Augmented CycleGAN: Learning Many-to-Many Mappings from Unpaired Data}

\begin{document}

\twocolumn[
\icmltitle{Augmented CycleGAN: Learning Many-to-Many Mappings \\ from Unpaired Data}

% It is OKAY to include author information, even for blind
% submissions: the style file will automatically remove it for you
% unless you've provided the [accepted] option to the icml2018
% package.

% List of affiliations: The first argument should be a (short)
% identifier you will use later to specify author affiliations
% Academic affiliations should list Department, University, City, Region, Country
% Industry affiliations should list Company, City, Region, Country

% You can specify symbols, otherwise they are numbered in order.
% Ideally, you should not use this facility. Affiliations will be numbered
% in order of appearance and this is the preferred way.

\icmlsetsymbol{workat}{\textdagger}

\begin{icmlauthorlist}
\icmlauthor{Amjad Almahairi}{mila,workat}
\icmlauthor{Sai Rajeswar}{mila}
\icmlauthor{Alessandro Sordoni}{msr}
\icmlauthor{Philip Bachman}{msr}
\icmlauthor{Aaron Courville}{mila,cifar}
\end{icmlauthorlist}

\icmlaffiliation{mila}{Montreal Institute for Learning Algorithms (MILA), Canada.}
\icmlaffiliation{msr}{Microsoft Research Montreal, Canada.}
\icmlaffiliation{cifar}{CIFAR Fellow. \textsuperscript{\textdagger}Work partly done at MSR Montreal}
% {Work done at Microsoft Research Montreal.}
\icmlcorrespondingauthor{Amjad Almahairi}{amjad.almahairi@umontreal.ca}

% You may provide any keywords that you
% find helpful for describing your paper; these are used to populate
% the "keywords" metadata in the PDF but will not be shown in the document
% \icmlkeywords{Machine Learning, ICML}

\vskip 0.3in
]

% this must go after the closing bracket ] following \twocolumn[ ...

% This command actually creates the footnote in the first column
% listing the affiliations and the copyright notice.
% The command takes one argument, which is text to display at the start of the footnote.
% The \icmlEqualContribution command is standard text for equal contribution.
% Remove it (just {}) if you do not need this facility.

\printAffiliationsAndNotice{} % leave blank if no need to mention equal contribution
% \printAffiliationsAndNotice{\icmlEqualContribution} % otherwise use the standard text.

\begin{abstract}
Learning inter-domain mappings from unpaired data can improve performance in structured prediction tasks, such as image segmentation, by reducing the need for paired data. CycleGAN was recently proposed for this problem, but critically assumes the underlying inter-domain mapping is approximately deterministic and one-to-one. This assumption renders the model ineffective for tasks requiring flexible, many-to-many mappings. We propose a new model, called Augmented CycleGAN, which learns many-to-many mappings between domains. We examine Augmented CycleGAN qualitatively and quantitatively on several image datasets.
\end{abstract}

\section{Introduction} \label{sec:introduction}

The problem of learning mappings between domains from unpaired data has recently received increasing attention, especially in the context of image-to-image translation~\citep{zhu2017unpaired, kim2017learning,liu2017unsupervised}. This problem is important because, in some cases, paired information may be scarce or otherwise difficult to obtain. For example, consider tasks like face transfiguration (male to female), where obtaining explicit pairs would be difficult as it would require artistic authoring. An effective unsupervised model may help when learning from relatively few paired examples, as compared to training strictly from the paired examples. Intuitively, forcing inter-domain mappings to be (approximately) invertible by a model of limited capacity acts as a strong regularizer.

% Given the relatively unlimited quantities of unlabeled data compared to labeled data, researchers have long considered unsupervised learning the true holy grail of machine learning. One interesting step in this direction is learning mappings across domains in an unsupervised way, i.e. using only unpaired data from each domain. These mappings can be thought of as learning simple analogy-making: they learn to make analogies between domains by leveraging the statistical structure in each of them. This is a small but important step on the road to extracting meaning from unstructured data, which has huge potential for solving hard structured prediction tasks.

% Unsupervised learning of mappings has received much recent attention, especially in the context of image-to-image translation~\citep{zhu2017unpaired, kim2017learning,yi2017dualgan,liu2017unsupervised}, where the tasks can be formulated as learning a conditional generative model which transforms an input image in one domain into an output image in the other domain.
%and machine translation~\citep{lample2017unsupervised}.
% The problem can be formulated as learning a conditional generative model which transforms an input in one domain into an output in the other domain. 
Motivated by the success of Generative Adversarial Networks (GANs) in image generation~\citep{goodfellow2014generative,radford2015unsupervised}, existing unsupervised mapping methods such as CycleGAN~\citep{zhu2017unpaired} learn a generator which produces images in one domain given images from the other. 
% While conditional GANs can be very effective for learning mappings that produce realistic images, a main challenge that remains is to keep a strong association between conditioned inputs and the generator's output; otherwise, multiple input images are mapped to the same output image, which known as the mode collapse problem in GANs.
Without the use of pairing information, there are many possible mappings that could be inferred. To reduce the space of the possible mappings, these models are typically trained with a \textit{cycle-consistency} constraint which enforces a strong connection across domains, by requiring that mapping an image from the source domain to the target domain and then back to source will result in the same starting image. This framework has been shown to learn convincing mappings across image domains and proved successful in a variety of related applications~\citep{tung2017adversarial,wolf2017unsupervised,hoffman2017cycada}.

\begin{figure}
\centering
  \begin{minipage}[b]{.2\textwidth}
  \subfloat[CycleGAN]{\label{fig:general_cgan}\includegraphics[width=0.9\textwidth]{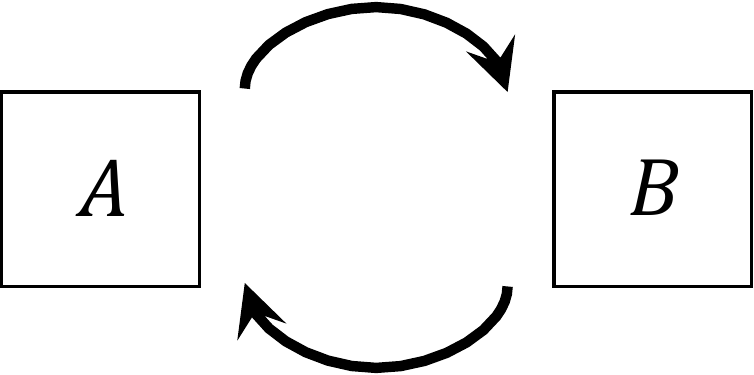}}
  \end{minipage}
\begin{minipage}[b]{.2\textwidth}
\subfloat[Augmented CycleGAN]{\label{fig:approach_general}\includegraphics[width=\textwidth]{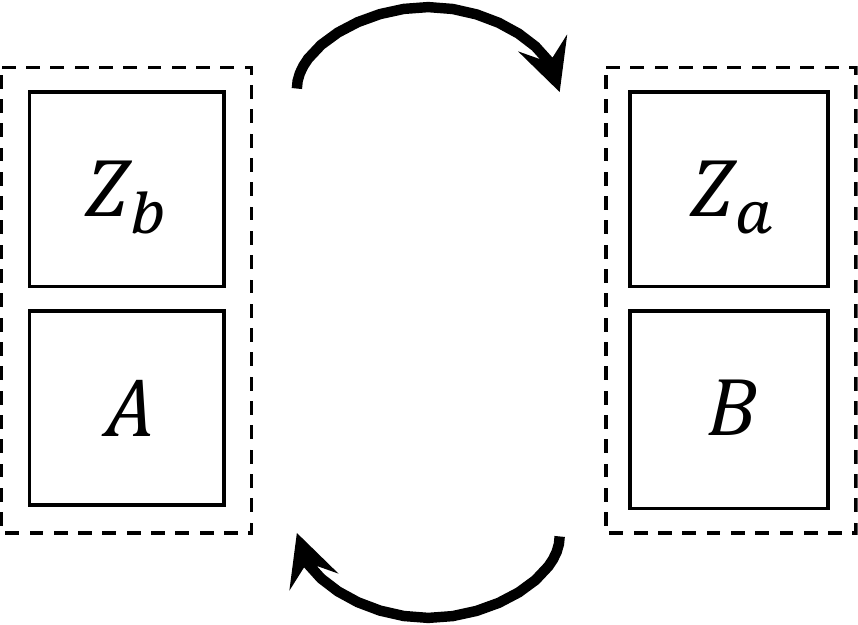}}
\end{minipage}
\caption{(a) Original CycleGAN model. (b) We propose to learn many-to-many mappings by cycling over the original domains augmented with auxiliary latent spaces. By marginalizing out auxiliary variables, we can model many-to-many mappings in between the domains.}
\end{figure}

%In order to avoid this degenerate behavior,~\citep{zhu2017unpaired} proposed the CycleGAN model that learns mappings in both directions simultaneously (i.e. $\A \mapsto \B$ and $\B \mapsto \A$), and satisfies a \textit{cycle-consistency} constraint which enforces a strong connection across domains by  that the mapping of an image from the source domain to the target domain and then back to source will result in the same original image we started with.

% This framework has been shown to learn very convincing mappings across image domains and proved successful in a variety of related applications~\citep{tung2017adversarial,wolf2017unsupervised,hoffman2017cycada}.
% \todo{Learning one-to-one mappings is a limitation of cyclegan}
One major limitation of CycleGAN is that it only learns one-to-one mappings,~i.e. the model associates each input image with a single output image. 
%\todo{Why is learning more flexible mappings important?}
%Should discuss multimodality in the sense of multiple data types speech and text, or image and text.
We believe that most relationships across domains are more complex, and better characterized as \textit{many-to-many}.
%From a theoretical perspective, a restricted, unimodal generative model provides a poor approximation of the true, multi-modal conditional. As a result, we empirically observe that CycleGAN learns to approximate these complex relationships with mappings that output a single and rather arbitrary output image given an input image. 
%\todo{reference an example of edges-to-shoes, and show that CycleGAN can produce a single shoe given an edge image}.
For example, consider mapping silhouettes of shoes to
images of shoes. While the mapping that CycleGAN learns can be superficially convincing (e.g.~it produces a single reasonable shoe with a particular style), we would like to learn a mapping that can capture diversity of the output (e.g.~produces multiple shoes with different styles). 
The limits of one-to-one mappings are more dramatic when the source domain and target domain substantially differ. For instance, it would be difficult to learn a CycleGAN model when the two domains are descriptive facial attributes and images of faces.

We propose a model for learning many-to-many mappings between domains from unpaired data. Specifically, we ``augment'' each domain with auxiliary latent variables and extend CycleGAN's training procedure to the augmented spaces. The mappings in our model take as input a sample from the source domain and a latent variable, and output both a sample in the target domain and a latent variable (Fig.~\ref{fig:approach_general}). The learned mappings are one-to-one in the augmented space, but many-to-many in the original domains after marginalizing over the latent variables.

Our contributions are as follows.
\begin{inlinelist}
\item We introduce the Augmented CycleGAN model for learning many-to-many mappings across domains in an unsupervised way.
\item We show that our model can learn mappings which produce a diverse set of outputs for each input.
% Amjad: I don't want reviewer think we do semi-supervised CIFAR10 or MNIST.
\item We show that our model can learn mappings across substantially different domains, and we apply it in a semi-supervised setting for mapping between faces and attributes with competitive results.
%\item We show that our model can be applied to semi-supervised settings, and show competitive results on several benchmarks.
\end{inlinelist}

\section{Unsupervised Learning of Mappings Between Domains}

\subsection{Problem Setting}
%NOTE: I think we should try to use learning conditional image generation instead of learning mapping. The former clarifies that the goal is to learn a conditional generative model, while the latter can be understood as mapping between existing items in the training datasets
Given two domains $\A$ and $\B$, we assume there exists a mapping, potentially many-to-many, between their elements. The objective is to recover this mapping using \textit{unpaired} samples from distributions $\pdata(a)$ and $\pdata(b)$ in each domain. 
%The mapping may be many-to-many.
%: $\{a^{(i)}\}_{i=1}^{N} \sim \pdata(a)$ and $\{b^{(i)}\}_{i=1}^{M} \sim \pdata(b)$, where $p_d(a)$ and $p_d(b)$ are data distributions.
This can be formulated as a conditional generative modeling task where we try to estimate the \textit{true conditionals} $p(a|b)$ and $p(b|a)$ using samples from the true marginals.
% Amjad: I'm not sure about how the following sentence..
An important assumption here is that elements in domains $\A$ and $\B$ are highly dependent; otherwise, it is unlikely that the model would uncover a meaningful relationship without any pairing information.

\subsection{CycleGAN Model}
The CycleGAN model~\citep{zhu2017unpaired} estimates these conditionals using two mappings $\F: \A \mapsto \B$ and $\G: \B \mapsto \A$, parameterized by neural networks, which satisfy the following constraints:
\begin{enumerate}
\item \textbf{Marginal matching}: The output of each mapping should match the empirical distribution of the target domain, when marginalized over the source domain.
\item \textbf{Cycle-consistency}: Mapping an element from one domain to the other, and then back, should produce a sample close to the original element.
\end{enumerate}

Marginal matching in CycleGAN is achieved using the generative adversarial networks framework (GAN)~\citep{goodfellow2014generative}. Mappings $\F$ and $\G$ are given by neural networks trained to fool adversarial discriminators $D_\B$ and $D_\A$, respectively. Enforcing marginal matching on target domain $\B$, marginalized over source domain $\A$, involves minimizing an adversarial objective with respect to $\F$:
\begin{equation}
\begin{aligned}
\label{eq:cgan_gan_loss}
  \loss_{\text{GAN}}^B(\F, D_\B) =& 
	\E_{b \sim \pdata(b)} \Big[ \log D_\B(b) \Big] +\\
    &\E_{a \sim \pdata(a)} \Big[ \log(1- D_\B(\F(a))) \Big],
\end{aligned}
\end{equation}
while the discriminator $D_\B$ is trained to maximize it. A similar adversarial loss $\loss_\text{GAN}^A(\G, D_\A)$ is defined for marginal matching in the reverse direction.

Cycle-consistency enforces that, when starting from a sample $a$ from $A$,
the reconstruction $a' = \G(\F(a))$ remains close to the original $a$.
For image domains, closeness between $a$ and  $a^{\prime}$ is typically measured with $L_1$ or $L_2$ norms. When using the $L_1$ norm, cycle-consistency starting from $\A$ can be formulated as:
\begin{equation}
\begin{aligned}
\label{eq:cgan_cycle_loss}
    &\loss_{\text{CYC}}^A(\F, \G) =& \E_{a \sim \pdata(a)} \big\| \G(\F(a)) - a \big\|_1.
%&\loss_{\text{CYC}}^B(\F, \G) =& \E_{b \sim \pdata(b)} \big\| \F(\G(b)) - b \big\|_1.
\end{aligned}
\end{equation}
And similarly for cycle-consistency starting from $\B$. The full CycleGAN objective is given by:
\begin{equation}
\begin{aligned}
\label{eq:cgan_full_loss}
%\min_{\F, \G} \max_{D_\A, D_\B} &\loss_{\text{GAN}}(\F, D_\B) +
% \loss_{\text{GAN}}(\G, D_\A) +\\ &\gamma \loss_{\text{Cycle}}(\F, \G),
&\loss^A_{\text{GAN}}(\G, D_\A) + \loss^B_{\text{GAN}}(\F, D_\B) \; + \\ & \gamma \loss_{\text{CYC}}^A(\F, \G) + \gamma \loss_{\text{CYC}}^B(\F, \G),
\end{aligned}
\end{equation}
where $\gamma$ is a hyper-parameter that balances between marginal matching and cycle-consistency.

The success of CycleGAN can be attributed to the complementary roles of marginal matching and cycle-consistency in its objective. Marginal matching encourages generating realistic samples in each domain. Cycle-consistency encourages a tight relationship between domains. It may also help prevent multiple items from one domain mapping to a single item from the other, analogous to the troublesome mode collapse in adversarial generators~\citep{li2017alice}.

\subsection{Limitations of CycleGAN} \label{sec:cgan_limitations}

% However, there are two problematic aspects that arise with this model.
% \paragraph{One-to-one mappings} 
A fundamental weakness of the CycleGAN model is that it learns deterministic mappings. In CycleGAN, and in other similar models~\citep{kim2017learning,yi2017dualgan}, the conditionals between domains correspond to delta functions: $\hat p(a | b) = \delta(\G(b))$ and $\hat p(b | a) = \delta(\F(a))$, and cycle-consistency forces the learned mappings to be inverses of each other. When faced with complex cross-domain relationships, this results in CycleGAN learning an arbitrary one-to-one mapping instead of capturing the true, structured conditional distribution more faithfully.
%\paragraph{Information Bottleneck}
Deterministic mappings are also an obstacle to optimizing cycle-consistency when the domains differ substantially in complexity, in which case mapping from one domain (e.g. class labels) to the other (e.g. real images) is generally one-to-many. Next, we discuss how to extend CycleGAN to capture more expressive relationships across domains.

\subsection{CycleGAN with Stochastic Mappings} \label{sec:stoch_cgan}
A straightforward approach for extending CycleGAN to model many-to-many relationships is to equip it with stochastic mappings between $A$ and $B$. Let $\Z$ be a latent space with a standard Gaussian prior $p(z)$ over its elements. We define mappings $\Fz: \A \times \Z \mapsto \B$ and $\Gz: \B \times \Z \mapsto \A$\footnote{To avoid clutter in notation, we reuse the same symbols of deterministic mappings.}. Each mapping takes as input a vector of auxiliary noise and a sample from the source domain, and generates a sample in the target domain. Therefore, by sampling different $z \sim p(z)$, we could in principle generate multiple $b$'s conditioned on the same $a$ and vice-versa.
We can write the marginal matching loss on domain $\B$ as:
\begin{equation}
\begin{aligned}
\label{eq:stoch_cgan_gan_loss}
  \loss^B_{\text{GAN}}(\Fz, D_\B) = \E_{b \sim \pdata(b)} &\Big[ \log D_\B(b) \Big] +\\
  \E_{ \substack{a \sim \pdata(a) \\ z \sim p(z)}} &\Big[ \log(1- D_\B(\Fz(a,z))) \Big].
\end{aligned}
\end{equation} 
Cycle-consistency starting from $\A$ is now given by:
\begin{equation}
\small
\begin{aligned}
\label{eq:stoch_cgan_cycle_loss}
& \loss_{\text{CYC}}^A(\Fz, \Gz) = \E_{\substack{a \sim \pdata(a) \\ z_1, z_2 \sim p(z)}} \big\| \Gz(\Fz(a, z_1), z_2) - a \big\|_1
\end{aligned}
\end{equation}
The full training loss is similar to the objective in Eqn.~\ref{eq:cgan_full_loss}. We refer to this model as \textit{Stochastic CycleGAN}.

%\begin{figure*}[t!]
%\centering
%\includegraphics[width=0.8\textwidth]{FIG/augmented_cyclegan.pdf}
%\caption{CycleGAN (left), Stochastic CycleGAN (center) and the proposed Augmented CycleGAN (right). Model components identified with color coding. Circles with double lines denote elements that have been generated by the model during the computational flow.}
%\label{fig:augmented_cycle_gan}
%\end{figure*}

\begin{figure*}[t!]
\centering
\includegraphics[width=0.95\textwidth]{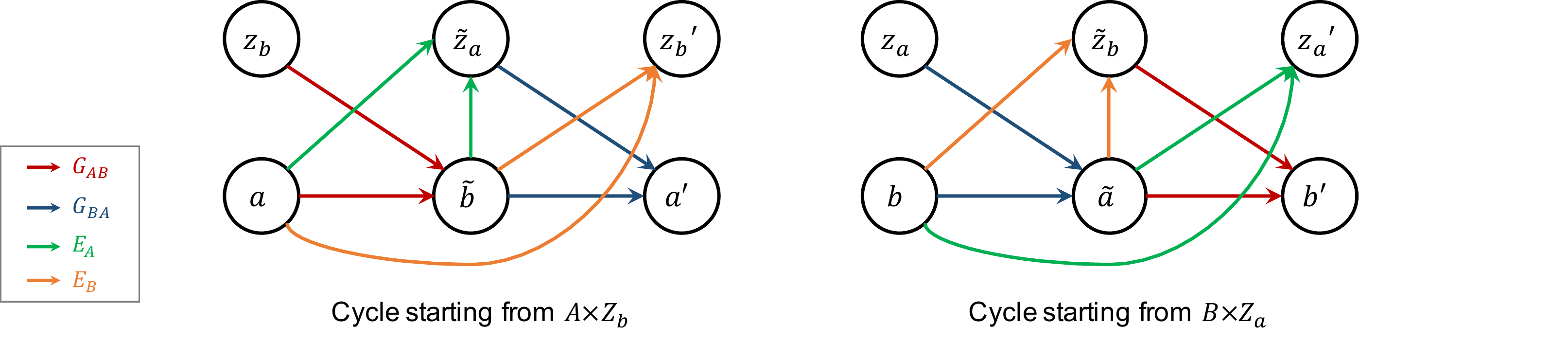}
\caption{Cycles starting from augmented spaces in Augmented CycleGAN. Model components identified with color coding.}
    \label{fig:augmented_cycle_gan}
\end{figure*}

In principle, stochastic mappings can model multi-modal conditionals, and hence generate a richer set of outputs than deterministic mappings.
However, Stochastic CycleGAN suffers from a fundamental flaw: the cycle-consistency in Eq.~\ref{eq:stoch_cgan_cycle_loss} encourages the mappings to ignore the latent $z$.
Specifically, the unimodality assumption implicit in the reconstruction error from Eq.~\ref{eq:stoch_cgan_cycle_loss} forces the mapping $\Gz$ to be many-to-one when cycling $A\rightarrow B \rightarrow A^{\prime}$, since any $b$ generated for a given $a$ must map to $a^{\prime} = \Gz(b, z) \approx a$, for all $z$.
For the cycle $B \rightarrow A \rightarrow B^{\prime}$, $\Fz$ is similarly forced to be many-to-one. The only way for to $\Gz$ and $\Fz$ to be both many-to-one and mutual inverses is if they collapse to being (roughly) one-to-one. We could possibly mitigate this degeneracy by introducing a VAE-like encoder and exchanging the $L_{1}$ error in Eq.~\ref{eq:stoch_cgan_cycle_loss} for a more complex variational bound on conditional log-likelihood. In the next section, we discuss an alternative approach to learning complex, stochastic mappings between domains.

%In fact, this model lacks an inference mechanism identifying latent codes that are most useful for producing a good reconstruction. In the following section, we propose an approach which addresses this shortcoming.

\section{Approach}
%In order to be able to optimize both marginal matching and cycle consistency when using stochastic mappings,
In order to learn many-to-many mappings across domains,
we propose to 
learn to map between pairs of items $(a, z_b) \in \A \times \Zb$ and $(b, z_a) \in \B \times \Za$, where $\Za$ and $\Zb$ are latent spaces that capture any missing information when transforming an element from $\A$ to $\B$, and vice-versa. For example, when generating a female face ($b \in \B$) which resembles a male face ($a \in \A$), the latent code $z_b \in \Zb$ can capture female face variations (e.g. hair length or style) independent from $a$. Similarly, $z_a \in \Za$ captures variations in a generated male face independent from the given female face. This approach can be described as learning mappings between \textit{augmented spaces} $\A \times \Zb$ and $\B \times \Za$ (Figure~\ref{fig:approach_general}); hence, we call it \textit{Augmented CycleGAN}.
By learning to map a pair $(a, z_b) \in \A \times \Zb$ to $(b, z_a) \in \B \times \Za$, we can
\begin{enumerate*}[label=(\roman*)]
\item learn a stochastic mapping from $a$ to multiple items in $\B$ by sampling different $z_b \in \Z_b$, and
 \item infer latent codes $z_a$ containing information about $a$ not captured in the generated $b$, which allows for doing proper reconstruction of $a$.
\end{enumerate*}
As a result, we are able to optimize both marginal matching and cycle consistency while using stochastic mappings.
We present details of our approach in the next sections.
\footnote{Our model captures many-to-many relationships because it captures both one-to-many and many-to-one: one item in A maps to many items in B, and many items in B map to one item in A (cycle). The same is true in the other direction.}

%In order to be able to optimize both marginal matching and cycle consistency when using stochastic mappings, we propose to augment domains $\A$ and $\B$ with latent spaces $\Zb$ and $\Za$ respectively and learn mappings between the augmented spaces $\A \times \Zb$ and $\B \times \Za$,~i.e. we learn to map between pairs of elements $(a, z_b) \in \A \times \Zb$ and $(b, z_a) \in \B \times \Za$.
%Samples from $\Zb$ ($\Za$) will be used jointly with an input element $a$ ($b$) to map to an element $b$ ($a$) in the other domain. This is similar to the model presented in the previous section. In contrast with the previous model however, we learn an additional mapping that produces an encoding $z_a$ ($z_b$), which carries the information about $a$ ($b$) that has been discarded when transforming an element from $\A$ to $\B$ ($B$ to $A$); this information is crucial to properly optimize the cycle-consistency cost on $a$ ($b$) in this stochastic setting (see Fig.~\ref{fig:augmented_cycle_gan}). From a general standpoint, this approach can be described as learning mappings over \textit{augmented spaces} $\A \times \Zb$ and $\B \times \Za$ (Fig.~\ref{fig:augmented_cycle_gan}); hence, we call it \textit{Augmented CycleGAN}.
%As an example, consider that $A$ and $B$ are male and female images of faces respectively. Then, samples $\Zb$ will encode typical features of female faces (e.g. hair length or lipstick) that may not be inferred from male faces. A similar argument can be done for the other direction.

\subsection{Augmented CycleGAN}
Our proposed model has four components. First, the two mappings $\Fz: \A \times \Zb \mapsto \B$ and $\Gz: \B \times \Za \mapsto \A$, which are the conditional generators of items in each domain. These models are similar to those used in Stochastic CycleGAN. We also have two encoders $\Ea:\A \times \B \mapsto \Za$ and $\Eb:\A \times \B \mapsto \Zb$, which enable optimization of cycle-consistency with stochastic, structured mappings.
All components are parameterized with neural networks -- see Fig.~\ref{fig:augmented_cycle_gan}.
We define mappings over augmented spaces in our model as follows. Let $\pz(z_a)$ and $\pz(z_b)$ be standard Gaussian priors over $\Za$ and $\Zb$, which are independent from $\pdata(b)$ and $\pdata(a)$. Given a pair $(a, z_b) \sim \pdata(a)\pz(z_b)$, 
%\footnote{We assume marginal independence between $\pdata(a)$ and $\pz(z_b)$, and $\pdata(b)$ and $\pz(z_a)$.} 
we generate a pair $(\tilde{b}, \tilde{z}_a)$ as follows: 
\begin{align}
\tilde{b} &= \Fz(a, z_b),\; \tilde{z}_a = \Ea(a, \tilde{b}).
\label{eq:aug_cgan_forward}
\end{align}
That is, we first generate a sample in domain $\B$, then we use it along with $a$ to generate latent code $\tilde{z}_a$. Note here that by sampling different $z_b \sim \pz(z_b)$, we can generate multiple $\tilde{b}$'s conditioned on the same $a$. In addition, given the pair $(a, \tilde{b})$, we can recover information about $a$ which is not captured in $\tilde{b}$, via $\tilde{z}_a$. %This is important for reconstructing $a$ as we will see in Sec~\ref{sec:aug_cgan_cycle}.
Similarly, given a pair $(b, z_a)\sim \pdata(b)\pz(z_a)$, we generate a pair $(\tilde{a}, \tilde{z}_b)$ as follows: 
\begin{align}
\tilde{a} &= \Gz(b, z_a),\; \tilde{z}_b = \Eb(b, \tilde{a}).
\end{align}
Learning in Augmented CycleGAN follows a similar approach to CycleGAN -- optimizing both marginal matching and cycle-consistency losses, albeit over augmented spaces. %We describe in the following these objectives in detail.

\paragraph{Marginal Matching Loss}
%The goal of marginal matching over $\B \times \Za$ is to ensure that the distribution of generated pairs $(\tilde{b}, \tilde{z}_a)$ matches the priors $(b,z_a) \sim \pdata(b)\pz(z_a)$.
We adopt an adversarial approach for marginal matching over $\B \times \Za$ where we use two independent discriminators $D_\B$ and $D_{\Za}$ to match generated pairs to real samples from the independent priors $\pdata(b)$ and $\pz(z_a)$, respectively. Marginal matching loss over $\B$ is defined as in Eqn~\ref{eq:stoch_cgan_gan_loss}. 
%as~\ref{eq:}
% Marginal matching loss over $\B$ is given by: 
% \begin{align}
% \label{eq:aug_cgan_gan_b}
%   \loss_{\text{GAN}}^B(\Fz, D_\B) = \E_{b \sim \pdata(b)} &\Big[ \log D_\B(b) \Big] +\nonumber \\
%   \E_{ \substack{a \sim \pdata(a) \\ z_b \sim p(z_b)}} &\Big[ \log(1- D_\B(\Fz(a,z_b))) \Big].
% \end{align}
Marginal matching over $\Za$ is given by:
\begin{align}
\label{eq:aug_cgan_gan_z}
  \loss_{\text{GAN}}^{\Za}(\Ea, \Fz, D_{\Za}) = &\E_{z_a \sim \pz(z_a)} \Big[ \log D_{\Za}(z_a) \Big] + \nonumber \\ 
  &\E_{ \substack{a \sim \pdata(a)\\ z_b \sim \pz(z_b)}} \Big[ \log(1- D_{\Za}(\tilde{z}_a)) \Big], %\nonumber \\
  %\tilde{z}_a =& \Ea(a, \tilde{b}), \nonumber \\
  %\tilde{b} =& \Gz(a, z_b).
\end{align}
where $\tilde{z}_a$ is defined by Eqn~\ref{eq:aug_cgan_forward}.
As in CycleGAN, the goal of marginal matching over $\B$ is to insure that generated samples $\tilde{b}$ are realistic. 
For latent codes $\tilde{z}_a$, marginal matching acts as a regularizer for the encoder, encouraging the marginalized encoding distribution to match a simple prior $\pz(z_a)$. This is similar to adversarial regularization of latent codes in adversarial autoencoders~\citep{makhzani2015adversarial}.
We define similar losses $\loss_{\text{GAN}}^A(\Gz, D_\A)$ and $\loss_{\text{GAN}}^{\Zb}(\Eb, \Gz, D_{\Zb})$ for marginal matching over $\A \times \Zb$.

\paragraph{Cycle Consistency Loss} % \label{sec:aug_cgan_cycle}
%The main motivation for performing cycle-consistency on augmented spaces is make sure that both source domain and latent space are used in generating samples.
We define two cycle-consistency constraints in Augmented CycleGAN starting from each of the two augmented spaces, as shown in Fig.~\ref{fig:augmented_cycle_gan}. 
In cycle-consistency starting from $\A \times \Zb$, we ensure that given a pair $(a,z_b) \sim \pdata(a)\pz(z_b)$, the model is able to produce a faithful reconstruction of it after being mapped to $(\tilde{b}, \tilde{z_a})$. This is achieved with two losses; first for reconstructing $a \sim \pdata(a)$:
\begin{align}
\label{eq:aug_cgan_fwd_cycle}
&\loss_{\text{CYC}}^A(\Fz, \Gz, \Ea) = \E_{\substack{a \sim \pdata(a)\\ z_b \sim \pz(z_b)}} \big\| a' - a \big\|_1 , \nonumber \\ %\; \text{where:} \nonumber \\
%\intertext{where:}
\tilde{b} &= \Fz(a, z_b), \;\tilde{z}_a = \Ea(a, \tilde{b}),\; a' = \Gz(\tilde{b}, \tilde{z}_a).
% a' &= \Gz(\tilde{b}, \tilde{z}_a), \nonumber \\
% \tilde{b} &= \Fz(a, z_b), \nonumber \\
% \tilde{z}_a &= \Ea(a, \tilde{b}).
\end{align}
The second is for reconstructing $z_b \sim \pz(z_b)$:
\begin{align}
\label{eq:aug_cgan_fwd_cycle_z}
\loss_{\text{CYC}}^{\Zb}(\Fz, \Eb) =& \E_{\substack{a \sim \pdata(a)\\ z_b \sim \pz(z_b)}} \big\| z_b' - z_b \big\|_1 , \nonumber \\ %\; \text{where:} \nonumber \\
%\intertext{where:}
z_b' = \Eb(a, \tilde{b}),& \;\;\; \tilde{b} = \Fz(a, z_b).
\end{align}
These reconstruction costs represent an autoregressive decomposition of the basic CycleGAN cycle-consistency cost from Eq.~\ref{eq:cgan_cycle_loss}, after extending it to the augmented domains. Specifically, we decompose the required reconstruction distribution $p(b, z_a | a, z_b)$ into the conditionals $p(b | a, z_b)$ and $p(z_a | a, z_b, b)$.

Just like in CycleGAN, the cycle loss in Eqn.~\ref{eq:aug_cgan_fwd_cycle} enforces the dependency of generated samples in $\B$ on samples of $\A$. Thanks to the encoder $\Ea$, the model is able to reconstruct $a$ because it can recover information loss in generated $\tilde{b}$ through $\tilde{z}_a$. 
On the other hand, the cycle loss in Eqn.~\ref{eq:aug_cgan_fwd_cycle_z} 
enforces the dependency of a generated sample $\tilde{b}$ on the given latent code $z_b$. In effect, it increases the mutual information between $z_b$ and $b$ conditioned on $a$, i.e. $I(b,z_b|a)$~\citep{chen2016infogan,li2017alice}.
% Likewise, we define backward cycle-consistency in Augmented CycleGAN with the following two losses:
% \begin{align}
% \label{eq:aug_cgan_bwd_cycle}
% \loss_{\overleftarrow{\text{Cycle}}}(\Fz, \Gz, \Eb) &= \E_{\substack{b \sim \pdata(b)\\ z_a \sim \pz(z_a)}} \Big[ \big\| b' - b \big\|_1  \Big], \nonumber \\ %\; \text{where:} \nonumber \\
% %\intertext{where:}
% b' &= \Fz(\tilde{a}, \tilde{z}_b), \nonumber \\
% \tilde{a} &= \Gz(b, z_a), \nonumber \\
% \tilde{z}_b &= \Eb(\tilde{a}, b).
% \end{align}
% \begin{align}
% \label{eq:aug_cgan_bwd_cycle_z}
% \loss_{\overleftarrow{\text{z-cycle}}}(\Gz, \Ea) &= \E_{\substack{b \sim \pdata(b)\\ z_a \sim \pz(z_a)}} \Big[ \big\| z_a' - z_a \big\|_1  \Big], \nonumber \\ %\; \text{where:} \nonumber \\
% %\intertext{where:}
% z_a' &= \Ea(\tilde{a}, b), \nonumber \\
% \tilde{a} &= \Gz(b, z_a).
% \end{align}

% \subsubsection{Full Training Loss}
Training Augmented CycleGAN in the direction $\A \times \Zb$ to $\B \times \Za$ is done by optimizing:
%\begin{equation}
%\begin{aligned}
%\label{eq:aug_cgan_full_loss}
%&\min_{\substack{ \Fz, \Ea\\ \Gz, \Eb}}
%\max_{D_\B, D_{\Za}}
%\loss_{\text{GAN}}(\Fz, D_\B) +
%\loss_{\text{GAN}}(\Ea, \Fz, D_{\Za}) +\\
%& \gamma \loss_{\text{Cycle}^a}(\Fz, \Gz, \Ea) + \gamma_z\loss_{\text{Cycle}^{z_b}}(\Fz, \Eb) 
%\end{aligned}
%\end{equation}
\begin{equation}
\begin{split}
\label{eq:aug_cgan_full_loss}
&\loss_{\text{GAN}}^B(D_\B, \Fz) + \loss_{\text{GAN}}^{z_a}(D_{\Za}, \Ea, \Fz) \, + \\
& \gamma_1 \loss_{\text{CYC}}^A(\Fz, \Gz, \Ea) + \gamma_2 \loss_{\text{CYC}}^{z_b}(\Fz, \Eb),
\end{split}
\end{equation}
where $\gamma_1$ and $\gamma_2$ are a hyper-parameters used to balance objectives. We define a similar objective for the direction going from $\B \times \Za$ to $\A \times \Zb$, and train the model on both objectives simultaneously.

\begin{figure}[t!]
\centering
\includegraphics[width=0.47\textwidth]{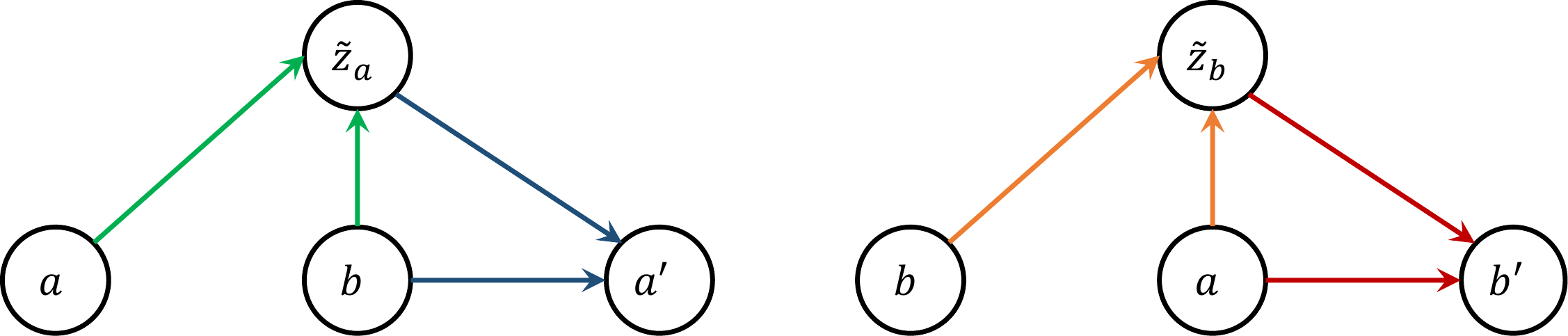}
\caption{Augmented CycleGAN when pairs $(a, b) \sim p_d(a, b)$ from the true joint distribution are observed. Instead of producing $\tilde{b}$ and $\tilde{a}$, the model uses samples from the joint distribution.
}
\label{fig:ss_augmented_cycle_gan}
\end{figure}

\subsection{Semi-supervised Learning with Augmented CycleGAN}
In cases where we have access to paired data, we can leverage it to train our model in a semi-supervised setting~ (Fig.~\ref{fig:ss_augmented_cycle_gan}).
Given pairs sampled from the true joint, i.e. $(a, b) \sim \pdata(a,b)$, we can define a supervision cost for the mapping $\Fz$ as follows:
\begin{align}
\label{eq:aug_cgan_fwd_sup}
\loss_{\text{SUP}}^A(\Gz, \Ea) =& \E_{(a,b) \sim \pdata(a,b)} \big\| \Gz(b, \tilde{z}_a) - a \big\|_1 , %\nonumber \\ %\; \text{where:} \nonumber \\
%\intertext{where:}
 %a' = \Gz(b, \tilde{z}_a).
% a' &= \Gz(\tilde{b}, \tilde{z}_a), \nonumber \\
% \tilde{b} &= \Fz(a, z_b), \nonumber \\
% \tilde{z}_a &= \Ea(a, \tilde{b}).
\end{align}
where $\tilde{z}_a = \Ea(a, b)$ infers a latent code which can produce $a$ given $b$ via $\Gz(b, \tilde{z}_a)$. We also apply an adversarial regularization cost on the encoder, in the form of Eqn.~\ref{eq:aug_cgan_gan_z}. Similar supervision and regularization costs can be defined for $\Gz$ and $\Eb$, respectively.
% This learning procedure resembles learning a conditional adversarial autoencoder.
% \begin{align}
% \label{eq:aug_cgan_gan_z}
%   \loss_{\text{GAN}}^{\Za}(\Ea, \Fz, D_{\Za}) = &\E_{z_a \sim \pz(z_a)} \Big[ \log D_{\Za} \Big] + \nonumber \\ 
%   &\E_{ (a,b) \sim \pdata(a,b)} \Big[ \log(1- D_{\Za}(\tilde{z}_a)) \Big].% \nonumber \\
%   %\tilde{z}_a =& \Ea(a, b). %\nonumber \\
%   %\tilde{b} =& \Gz(a, z_b).
% \end{align}
% %where $\tilde{z}_a = \Ea(a,b)$
% \input{new-approach}
% \input{other-approach}

\subsection{Modeling Stochastic Mappings}
% \begin{enumerate}
% \item injecting noise in high-level feature maps
% \item CIN instead of concatenation
% \end{enumerate}
We note here some design choices that we found important for training our stochastic mappings. We discuss architectural and training details further in Sec.~\ref{sec:experiments}.
% We adopt standard multi-layer neural network architectures for modeling components of our model. For image translation tasks, $\F$ and $\G$ are conditional image generators adapted from~\citep{isola2017image} and \citep{zhu2017unpaired} to handle additional the latent codes, while for translation across attributes and images we use standard DCGAN~\citep{radford2015unsupervised} architectures.
In order to allow the latent codes to capture diversity in generated samples, we found it important to inject latent codes to layers of the network which are closer to the inputs. This allows the injected codes to be processed with a larger number of remaining layers and therefore capture high-level variations of the output, as opposed to small pixel-level variations.
We also found that Conditional Normalization (CN)~\citep{dumoulin2016learned,perez2017film} for conditioning layers can be more effective than concatenation, which is more commonly used~\citep{radford2015unsupervised,zhu2017toward}.
The basic idea of CN is to replace parameters of affine transformations in normalization layers~\citep{ioffe2015batch} of a neural network with a learned function of the conditioning information. We apply CN by learning two linear functions $f$ and $g$ which take a latent code $z$ as input and output scale and shift parameters of normalization layers in intermediate layers, i.e. $\gamma = f(z)$ and $\beta = g(z)$. When activations are normalized over spatial dimensions only, we get Conditional Instance Normalization (CIN), and when they are also normalized over batch dimension, we get Conditional Batch Normalization (CBN).

%One of the key architectural choices in our model is designing stochastic mappings $\Fz$ and $\Gz$.
%When domain $\A$ is a high dimensional space, e.g. images, the mapping $\Fz$ combines a low-dimensional latent code ($z_b \in \mathbb{R}^n$) with high dimensional input ($a \in \mathbb{R}^{H\times W\times 3}$) to produce an output in $\B$. 

% \subsubsection{An Autoencoder Perspective on Augmented CycleGAN}
% It's also two adversarial autoencoders but with auxiliary variables.

%\newpage % put this just to shut up sharelatex.. will remove it later
% Amjad: I commented this out because I believe we can cite all papers in the introduction. Nothing really is added here.

\section{Related Work}
%\paragraph{Unsupervised Learning of Mappings}
There has been a surge of interest recently in unsupervised learning of cross-domain mappings, especially for image translation tasks. Previous attempts for image-to-image translation have unanimously relied on GANs to learn mappings that produce compelling images. In order to constrain learned mappings, some methods have relied on cycle-consistency based constraints similar to CycleGAN~\citep{kim2017learning, yi2017dualgan, royer2017xgan}, while others relied on weight sharing constraints~\cite{liu2016coupled,liu2017unsupervised}. However, the focus in all of these methods was on learning conditional image generators that produce single output images given the input image.
Notably,~\citet{liu2015faceattributes} propose to map inputs from both domains into a shared latent space. This approach may constrain too much the space of learnable mappings, for example in cases where the domains differ substantially (class labels and images).
%Most recently, \citet{royer2017xgan} propose a method for unpaired image translation which holds promise for 

Unsupervised learning of mappings have also been addressed recently in language translation, especially for machine translation~\citep{lample2017unsupervised} and text style transfer~\citep{shen2017style}. 
These methods also rely on some notion of cycle-consistency over domains in order to constrain the learned mappings. 
They rely heavily on the power of the RNN-based decoders to capture complex relationships across domains while we propose to use auxiliary latent variables. The two approaches may be synergistic, as it was recently suggested in~\cite{gulrajani2016pixelvae}.
% \paragraph{Weakly Supervised Learning of Mappings}
% one-sided \citep{benaim2017one}, cross-modal-scene-networks \citep{aytar2017cross}, DTN \citep{taigman2016unsupervised}
%\paragraph{Supervised Learning of Multi-Modal Mappings}

Recently, \citet{zhu2017toward} proposed the BiCycleGAN model for learning multi-modal mappings but in fully supervised setting. This model extends the pix2pix framework in~\cite{isola2017image} by learning a stochastic mapping from the source to the target, and shows interesting diversity in the generated samples. 
Several modeling choices in BiCycleGAN resemble our proposed model, including the use of stochastic mappings and an encoder to handle multi-modal targets. However, our approach focuses on unsupervised many-to-many mappings, which allows it to handle domains with no or very little paired data. 
%While their approach bears strong resemblance to the supervised costs defined for semi-supervised settings in Augmented CycleGAN, our model tackles a 
%While the way mappings in BiCycleGAN are learned in a similar way to the semi-supervised approach in Augmented CycleGAN, their model cannot  

\section{Experiments} \label{sec:experiments}

\subsection{Edges-to-Photos} \label{sec:e2s_experiments}
We first study a one-to-many image translation task between edges (domain $\A$) and photos of shoes (domain $\B$).\footnote{ Public code available at: \url{https://github.com/aalmah/augmented_cyclegan}}
%The relationship between (domain $\A$) and shoes (domain $\B$) is one-to-many, as there many possible images of shoes that can be associated with a given edge image.
Training data is composed of almost 50K shoe images with corresponding edges~\citep{yu2014fine,zhu2016generative,isola2017image}, but as in previous approaches (e.g. \citep{kim2017learning}), we assume no pairing information while training unsupervised models. Stochastic mappings in our Augmented CycleGAN (AugCGAN) model are based on ResNet conditional image generators of~\citep{zhu2017unpaired}, where we inject noise with CIN to all intermediate layers.
%~\footnote{Architecture and training details of this experiment are provided in Sec.\todo{name section.}}
%of $\Fz$, and use a deterministic mapping from shoes to edges, i.e. $\Gz=\G$, since the relationship is one-to-many. 
As baselines, we train: CycleGAN, Stochastic CycleGAN (StochCGAN) and Triangle-GAN ($\Delta$-GAN) of~\citep{gan2017triangle} which share the same architectures and training procedure for fair comparison.~\footnote{$\Delta$-GAN architecture differs only in the two discriminators, which match conditionals/joints instead of marginals.}

\begin{table*}[t!]
\begin{minipage}[b]{.35\linewidth}
\centering
\includegraphics[width=\linewidth]{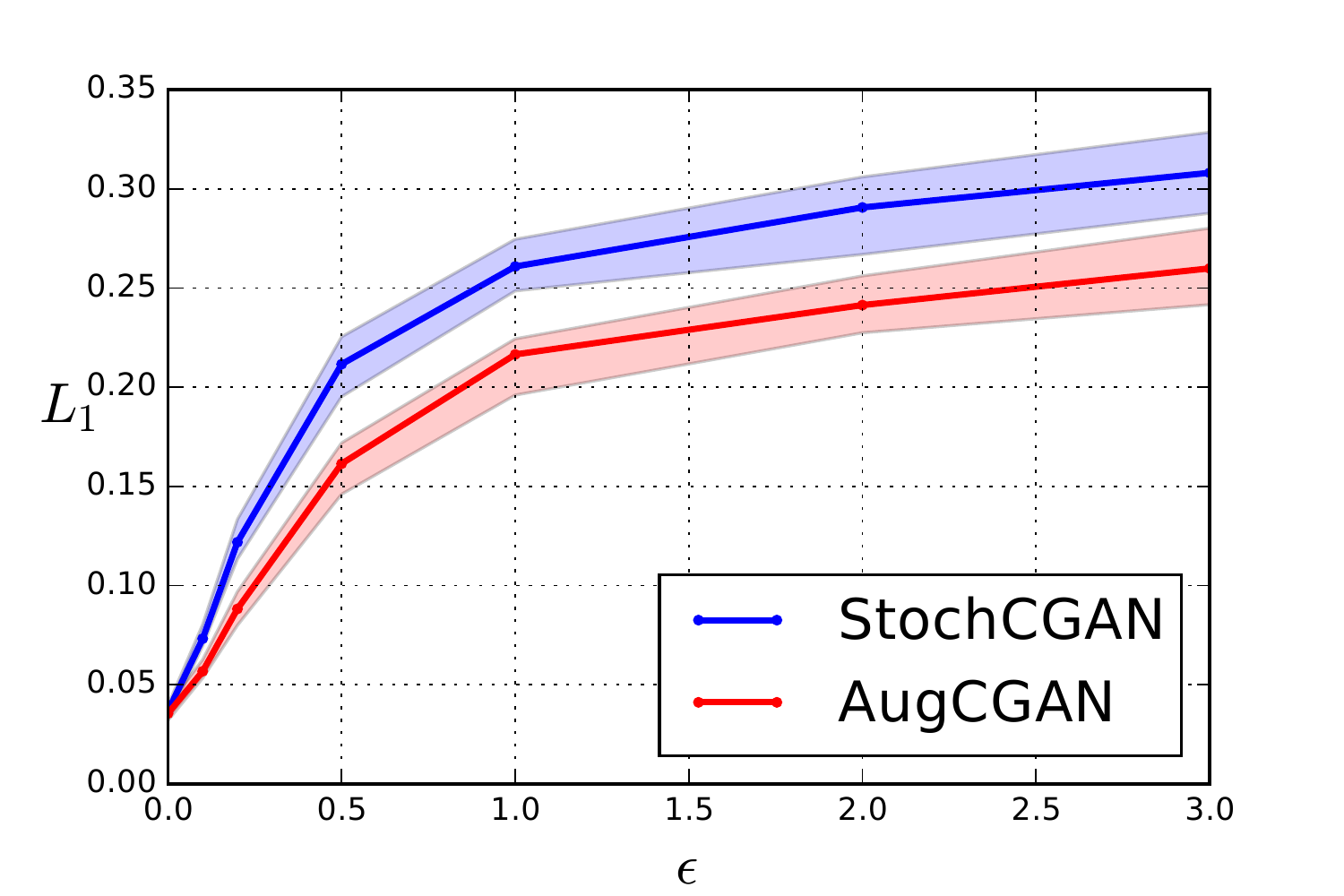}
\captionof{figure}{Shoes reconstruction error given a generated edge as a function of the Gaussian noise $\epsilon$ injected in the generated edge.}
    \label{fig:noise_sensitivity}
\end{minipage}
\hfill
\begin{minipage}[b]{.3\linewidth}
\centering
\begin{tabular}{lcc}
    \toprule
    Model (Paired \%)    &   Avg. $L_1$\\
    \midrule
    CycleGAN  (0\%)& 0.1837 \\
    StochCGAN (0\%) & 0.0794 \\
    $\Delta$-GAN$^\dagger$ (10\%)  & 0.0748 \\
    \midrule
    AugCGAN  (0\%) & \textbf{0.0698} \\
	AugCGAN (10\%)  & \textbf{0.0562} \\
    \bottomrule
  \end{tabular}
  \caption{\label{tab:shoe_recons}Reconstruction error for shoes given edges in the test set. $^\dagger$Same architecture as our model.}
\end{minipage}
% \begin{minipage}[b]{.3\linewidth}
% \centering
% \begin{tabular}{lcc}
%     \toprule
%     Model (Paired Data)    &  BPP &  $L_1$\\
%     \midrule
%     CycleGAN  (0\%)& 6.77 & 0.1837 \\
%     StochCGAN (0\%) & 5.53 & 0.0794 \\
%     AugCGAN  (0\%) & 5.43 & 0.0698 \\
%     $\Delta$-GAN$^\dagger$ (10\%)   & 5.48 & 0.0748 \\
% 	AugCGAN (10\%)  & 5.13 & 0.0562 \\
%     \bottomrule
%   \end{tabular}
%   \caption{Shoe likelihood. $^\dagger$ Our implementation.}
% \end{minipage}
\hfill
\begin{minipage}[b]{.3\linewidth}
\centering
\begin{tabular}{lc}
    \toprule
    Model (Paired \%)    & MSE\\
    \midrule
    $\Delta$-GAN$^\star$ (10\%)  & 0.0102 \\
    $\Delta$-GAN$^\dagger$ (10\%)  & 0.0096 \\
    $\Delta$-GAN$^\star$ (20\%) & 0.0092 \\
    \midrule
    AugCGAN (0\%)   & \textbf{0.0079} \\
    AugCGAN (10\%)  & \textbf{0.0052} \\
    \bottomrule
  \end{tabular}
  \caption{MSE on edges given shoes in the test set. $^\star$ From~\citep{gan2017triangle}. $^\dagger$Same architecture as our model.}
   \label{tab:edge_recons}
\end{minipage} 
\end{table*}

\begin{table*}[t]
\centering
\begin{minipage}[t]{.45\textwidth}
%  \begin{minipage}[b]{.22\textwidth}
  \subfloat[AugCGAN]
  {\includegraphics[width=0.45\textwidth]{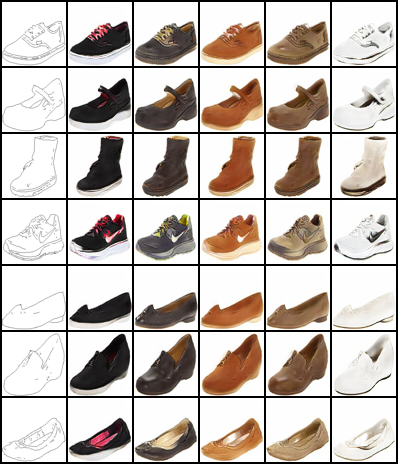}}
  \hfill
%  \end{minipage}
%\begin{minipage}[b]{.22\textwidth}
  \subfloat[StochCGAN]
  {\includegraphics[width=0.45\textwidth]{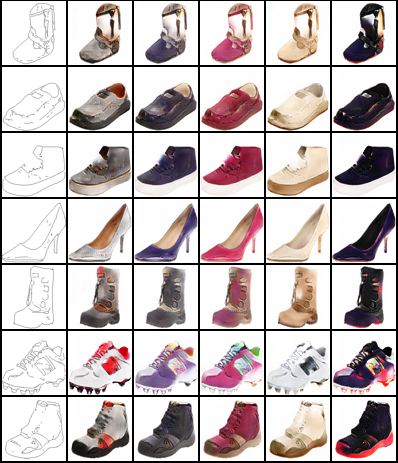}}
%  \end{minipage}
\captionof{figure}{Given an edge from the data distribution (leftmost column), we generate shoes by sampling five $z_b \sim \pz(z_b)$. Models generate diverse shoes when edges are from the data distribution.}
\label{fig:e2s_multi}
\end{minipage}
\hfill
\begin{minipage}[t]{.49\textwidth}
  \subfloat[AugCGAN]
  {\includegraphics[width=0.45\textwidth]{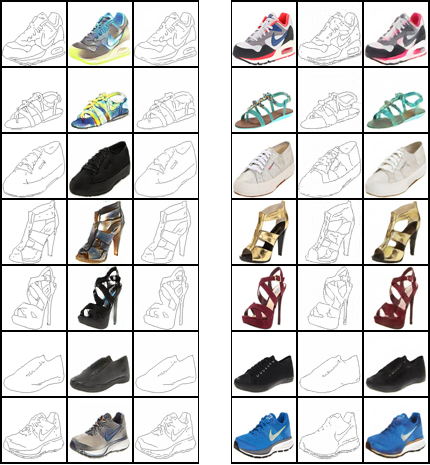}}
  \hfill
  \subfloat[StochCGAN]
  {\includegraphics[width=0.45\textwidth]{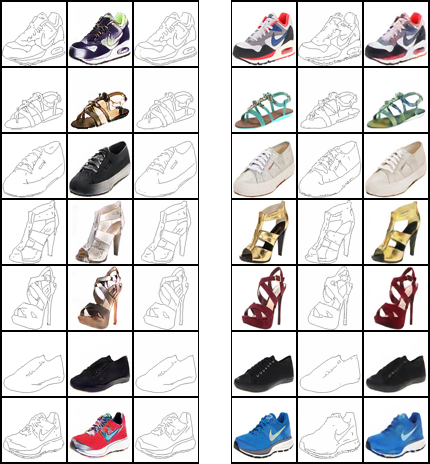}}
\captionof{figure}{Cycles from both models starting from a real edge and a real shoe (left and right respectively in each subfigure). The ability for StochCGAN to reconstruct shoes is surprising and is due to the ``steganography'' effect (see text).}
\label{fig:e2s_cycle}
\end{minipage}
\end{table*}

\paragraph{Quantitative Results}

First, we evaluate conditionals learned by each model by measuring the ability of the model of generating a specific edge-shoe pair from a test set. We follow the same evaluation methodology adopted in~\cite{metz2016unrolled,xiang2017effects}, which opt for an inference-via-optimization approach to estimate the reconstruction error of a specific shoe given an edge. Specifically, given a trained model with mapping $\Fz$ and an edge-shoe pair $(a,b)$ in the test set, we solve the optimization task $z_b^* = \argmin_{z_b} \| \Fz(a,z_b) - b \|_1$ and compute reconstruction error $\| \Fz(a,z_b^*)-b\|_1$. Optimization is done with RMSProp as in~\citep{xiang2017effects}. We show the average errors over a predefined test set of 200 samples in Table~\ref{tab:shoe_recons} for: AugCGAN (unsupervised and semi-supervised with 10\% paired data), unsupervised CycleGAN and StochCGAN, and a semi-supervised $\Delta$-GAN, all sharing the same architecture. Our unsupervised AugCGAN model outperforms all baselines including semi-supervised $\Delta$-GAN, which indicates that reconstruction-based cycle-consistency is more effective in learning conditionals than the adversarial approach of $\Delta$-GAN. 
As expected, adding 10\% supervision to AugCGAN improves shoe predictions further. 
In addition, we evaluate edge predictions given real shoes from test set as well. We report mean squared error (MSE) similar to~\citep{gan2017triangle}, where we normalize over all edge pixels. The $\Delta$-GAN model with our architecture outperforms the one reported in~\citep{gan2017triangle}, but is outperformed by our unsupervised AugCGAN model. Again, adding 10\% supervision to AugCGAN reduces MSE even further.

\paragraph{Qualitative Results} 
We qualitatively compare the mappings learned by our model AugCGAN and StochCGAN. Fig.~\ref{fig:e2s_cycle} shows generated images of shoes given an edge $a \sim \pdata(a)$ (row) and $z_b \sim p(z_b)$ (column) from both model, and Fig.~\ref{fig:e2s_multi} shows cycles starting from edges and shoes. Note that here the edges are sampled from the data distribution and not produced by the learnt stochastic mapping $\G$.
In this case, both models can
\begin{inlinelist}
\item generate diverse set of shoes with color variations mostly defined by $z_b$, and
\item perform reconstructions of both edges and shoes.
\end{inlinelist}

\begin{figure}[t!]
\centering
  \begin{minipage}[b]{.22\textwidth}
  %\subfloat[AugCGAN]{\includegraphics[width=0.9\textwidth]{FIG/edges2shoes/aug_cgan_short/cycle_B_multi_1.png}}
  \subfloat[AugCGAN]{\includegraphics[width=0.9\textwidth]{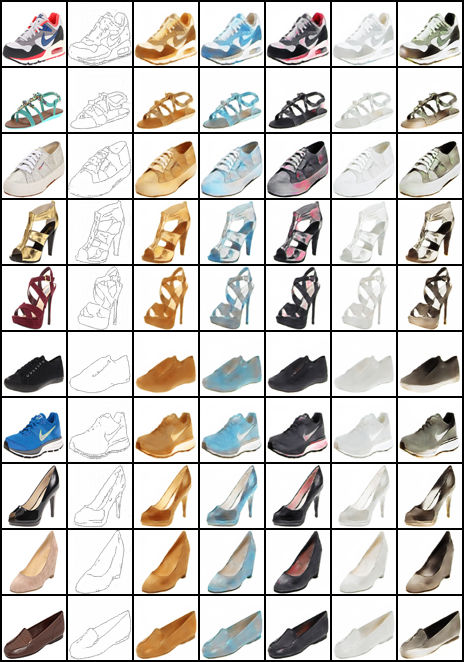}}
  \end{minipage}
\begin{minipage}[b]{.22\textwidth}
  %\subfloat[StochCGAN]{\includegraphics[width=0.9\textwidth]{FIG/edges2shoes/stoch_cgan_short/cycle_B_multi_0.png}}
  \subfloat[StochCGAN]{\includegraphics[width=0.9\textwidth]{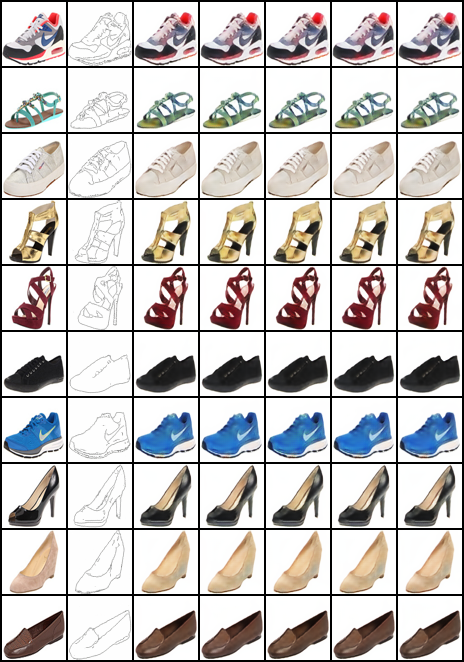}}
  \end{minipage}
\caption{Given a shoe from the data distribution (leftmost column), we generate an edge using the model (second column). Then, we generate shoes by sampling five  $z_b \sim p(z_b)$. When edges are generated by the model, StochCGAN collapses to a single mode of the shoes distribution and generate the same shoe.}
\label{fig:e2s_cycle_B_multi}
\end{figure}

\begin{figure}[h]
\centering
  \begin{minipage}[b]{.23\textwidth}
  \subfloat[AugCGAN]{\includegraphics[width=0.95\textwidth]{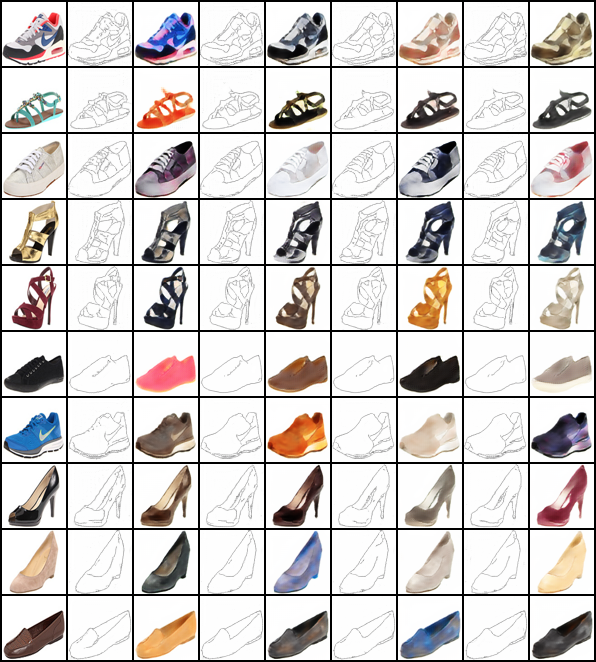}}
  \end{minipage}
\begin{minipage}[b]{.23\textwidth}
  \subfloat[StochCGAN]{\includegraphics[width=0.95\textwidth]{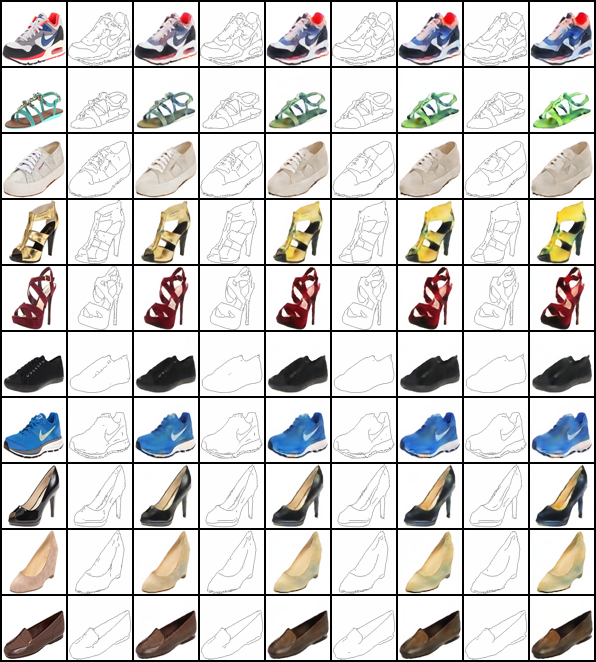}}
  \end{minipage}
\caption{We perform multiple generation cycles from the model by applying the learned mappings in turn. StochCGAN cycles collapse to the same shoe at each step which indicates that it doesn't capture the data distribution.}
\label{fig:e2s_multi_cycle}
\end{figure}

While we expect our model to achieve these results, the fact that StochCGAN can reconstruct shoes perfectly without an inference model may seem at first surprising. However, this can be explained by the ``steganography" behavior of CycleGAN~\citep{chu2017cyclegan}: the model hides in the generated edge $\tilde{a}$ imperceptible information about a given shoe $b$ (e.g. its color), in order to satisfy cycle-consistency without being penalized by the discriminator on $\A$. 
A good model of the true conditionals $p(b|a)$, $p(a|b)$ should reproduce the hidden joint distribution and consequently the marginals by alternatively sampling from conditionals. Therefore, we examine the behavior of the models when edges are generated from the model itself (instead of the empirical data distribution). In Fig.~\ref{fig:e2s_cycle_B_multi}, we plot
multiple generated shoes given an edge generated by the model,~i.e. $\tilde{a}$, and 5 different $z_b$ sampled from $p(z_b)$. In StochCGAN, the mapping $\Gz(\tilde{a}, z_b)$ collapses to a deterministic function generating a single shoe for every $z_b$. This distinction between behaviour on real and synthetic data is undesirable, e.g. regularization benefits of using unpaired data may be reduced if the model slips into this “regime switching” style. In AugCGAN, on the other hand, the mapping seem to closely capture the diversity in the conditional distribution of shoes given edges. 
% does not collapse and generate a diverse set of reconstructed shoes sharing the same shape of the original shoe but with colors captured by $z_b$.
Furthermore, in Fig.~\ref{fig:e2s_multi_cycle}, we run a Markov chain by generating from the learned mappings multiple times, starting from a real shoe. Again AugCGAN produces diverse samples while StochCGAN seems to collapse to a single mode.
%Fig.~\ref{fig:e2s_multi_cycle} provides another comparison between AugCGAN and StochCGAN multiple cycles through each model, where start from a real shoe and reconstruct it with $z_b \sim p(z_b)$ in every cycle step.

We  investigate ``steganography'' behavior in both AugCGAN and StochCGAN using a similar approach to~\citep{chu2017cyclegan}, where we corrupt generated edges with noise sampled from $\mathcal{N}(0,\epsilon^2)$, and compute reconstruction error of shoes. Fig.~\ref{fig:noise_sensitivity} shows $L_1$ reconstruction error as we increase $\epsilon$. AugCGAN seems more robust to corruption of edges than in StochCGAN, which confirms that information is being stored in the latent codes instead of being completely hidden in generated edges.

% \begin{table*}[t]
% \begin{minipage}[b]{.4\linewidth}
% \centering
% \includegraphics[width=\linewidth]{FIG/edges2shoes/noise_sensitivity.pdf}
% \captionof{figure}{Sensitivity to corruption.}
%     \label{fig:noise_sensitivity}
% \end{minipage}
% \hfill
% \end{table*}

% \begin{figure}
% \centering
%   \begin{minipage}[b]{.2\textwidth}
%   \subfloat[AugCGAN]
%   {\includegraphics[width=0.9\textwidth]{FIG/edges2shoes/aug_cgan_short/multi_1.png}}
%   %{\label{fig:e2s_augcgan_multi}\includegraphics[width=0.9\textwidth,trim={0 11.65cm 0 0},clip]{FIG/edges2shoes/aug_cgan/multi_1.png}}
%   \end{minipage}
% \begin{minipage}[b]{.2\textwidth}
%   \subfloat[StochCGAN]{\includegraphics[width=0.9\textwidth]{FIG/edges2shoes/stoch_cgan_short/multi_2.png}}
%   \end{minipage}
% \caption{Shoe samples given edges and $z_b \sim \pz(z_b)$ shared across rows. Both models generate diverse shoes.}
% \label{fig:e2s_multi}
% \end{figure}

% \begin{figure}
% \centering
%   \begin{minipage}[b]{.22\textwidth}
%   \subfloat[AugCGAN]{\includegraphics[width=0.9\textwidth]{FIG/edges2shoes/aug_cgan_short/cycle_short.png}}
%   \end{minipage}
% \begin{minipage}[b]{.22\textwidth}
%   \subfloat[StochCGAN]{\includegraphics[width=0.9\textwidth]{FIG/edges2shoes/stoch_cgan_short/cycle_short.png}}
%   \end{minipage}
% \caption{Forward and Backward Cycles. Both models can perform good reconstructions.}
% \label{fig:e2s_cycle}
% \end{figure}
\setlength{\belowcaptionskip}{-10pt}

\subsection{Male-to-Female} \label{sec:m2f_experiments}

\begin{figure}[h]
\centering
\includegraphics[width=.45\textwidth]{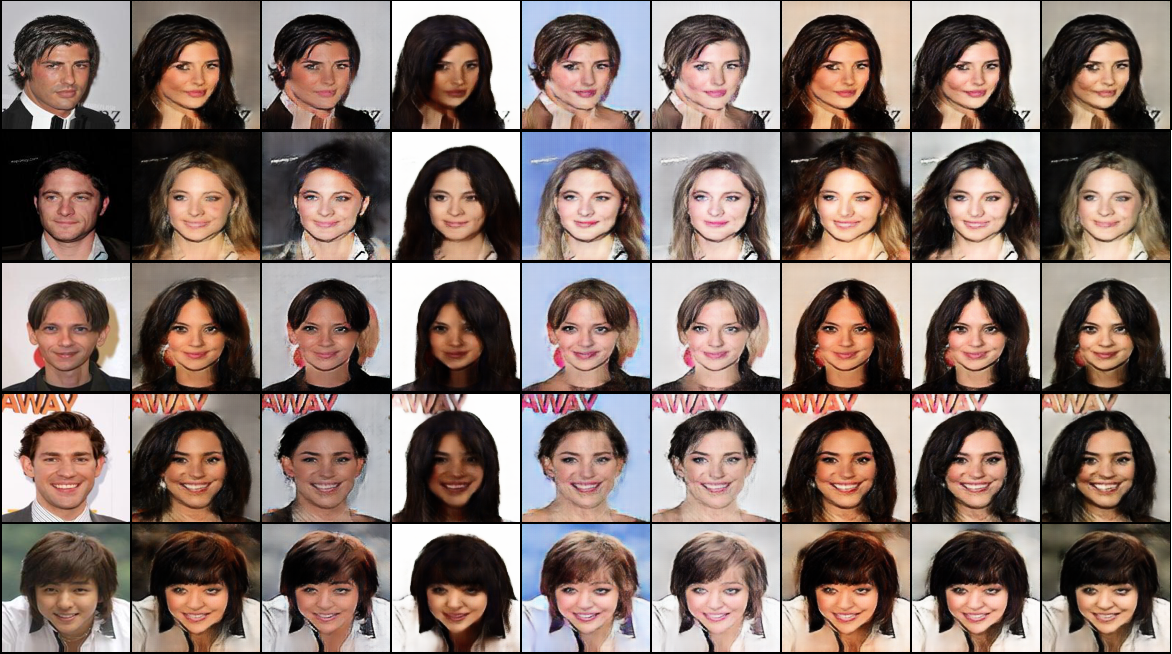}
\caption{Given a male face from the data distribution (leftmost column), we generate 8, 128$\times$128 female faces with AugCGAN by sampling $z_b \sim p(z_b)$.}
\label{fig:m2f_128}
\end{figure}

\begin{figure*}[th!]
\centering
  \begin{minipage}[b]{\textwidth}
  \centering
  \subfloat[AugCGAN Female-to-Male]{\includegraphics[width=.24\textwidth]{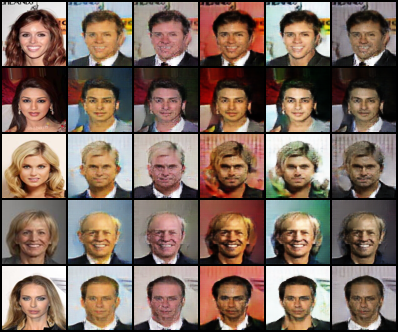}}
  \hfill
  \subfloat[AugCGAN Male-to-Female]{\includegraphics[width=.24\textwidth]{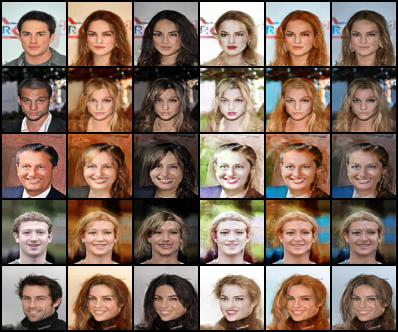}}
  \hfill
  \subfloat[$z$ in last 3 layers only]{\includegraphics[width=.24\textwidth]{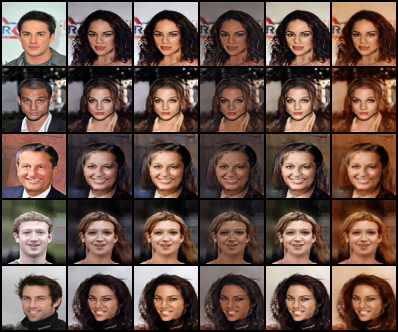}}
  \hfill
  \subfloat[StochCGAN]{\includegraphics[width=.24\textwidth]{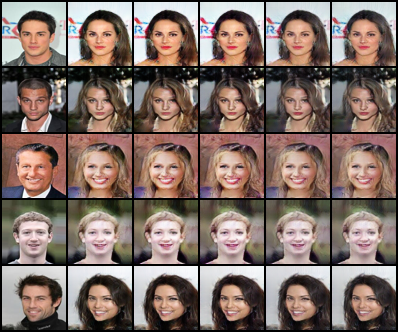}}
  \caption{Generated $64\times 64$ faces given a real face image from the other domain and multiple latent codes from prior.}
  \label{fig:m2f_comparison}
  \end{minipage}
\end{figure*}

% \begin{figure}

%   \begin{minipage}[b]{0.23\textwidth}
 
%   \subfloat[AugCGAN]{\includegraphics[width=.99\textwidth]{FIG/male2female/aug_cgan/multi_A_B.png}}
%   \end{minipage}
% %   \includegraphics[width=0.47\textwidth]{FIG/male2female/aug_cgan/multi_B_A.png}
%  \begin{minipage}[b]{0.23\textwidth}
%   \subfloat[StochCGAN]{\includegraphics[width=.99\textwidth]{FIG/male2female/stoch_cgan/multi_A_B.png}}
%   \end{minipage}
%   \caption{Comparision between(a) AugCGAN with CIN  (b) StochCGAN with CIN.}
%   \label{fig:m2f_comparison}
 
% \end{figure}

We study another image translation task of translating between male and female faces. Data is based on CelebA dataset~\cite{liu2015faceattributes} where we split it into two separate domains using provided attributes. Several key features distinguish this task from other image-translation tasks:
\begin{inlinelist}
\item there is no predefined correspondence in real data of each domain,
\item the relationship is many-to-many between domains, as we can map a male to female face, and vice-versa, in many possible ways, and
\item capturing realistic variations in generated faces requires transformations that go beyond simple color and texture changes.
\end{inlinelist}
The architecture of stochastic mappings are based on U-NET conditional image generators of~\cite{isola2017image}, and again with noise injected to all intermediate layers.
Fig.~\ref{fig:m2f_128} shows results of applying our model to this task on $128 \times 128$ resolution CelebA images.
%\footnote{The detailed experimental setup is provided in the appendix~\ref{sec:m2f_exp_details}}. 
We can see that our model depicts meaningful variations in generated faces without compromising their realistic appearance. In Fig.~\ref{fig:m2f_comparison} we show $64 \times 64$ generated samples in both domains from our model ((a) and (b)), and compare them to both: (c) our model but with noise injected noise only in last 3 layers of the $\Fz$'s network, and (d) StochCGAN with the same architecture.
We can see that in Fig.~\ref{fig:m2f_comparison}-(c) variations are very limited, which highlights the importance of processing latent code with multiple layers. StochCGAN in this task produces almost no variations at all, which highlights the importance of proper optimization of cycle-consistency for capturing meaningful variations.
We verify these results quantitatively using LPIPS distance~\cite{zhang2018perceptual}, where we average distance between 1000 pairs of generated female faces (10 random pairs from 100 male faces). AugCGAN (Fig.~\ref{fig:m2f_comparison}-(b)) achieves highest LPIPS diversity score with 0.108 $\pm$ 0.003, while AugCGAN with $z$ in low-level layers (Fig.~\ref{fig:m2f_comparison}-(c)) gets 0.059 +/- 0.001, and finally StochCGAN (Fig.~\ref{fig:m2f_comparison}-(d)) gets 0.008 +/- 0.000, i.e. severe mode collapse.

% \begin{table}[!t]
%   \centering
%   \small
%   \begin{tabular}{lccc}
%     \toprule
%     		       &  AugCGAN  & AugCGAN-low-$z$ & StochCGAN  \\
%     \midrule
%     \textbf{LPIPS}    & 0.108 $\pm$ 0.003 & 0.059 +/- 0.001 & 0.008 +/- 0.000 \\
%     \bottomrule
%   \end{tabular}
%   \caption{Diversity of generated female faces from models of Fig.~\ref{fig:m2f_comparison} (b), (c), and (d) using LPIPS~\cite{zhang2018perceptual}.}
%   \label{tab:lpips}
% \end{table}

\begin{figure}[t!]
\centering
\includegraphics[scale=0.45]{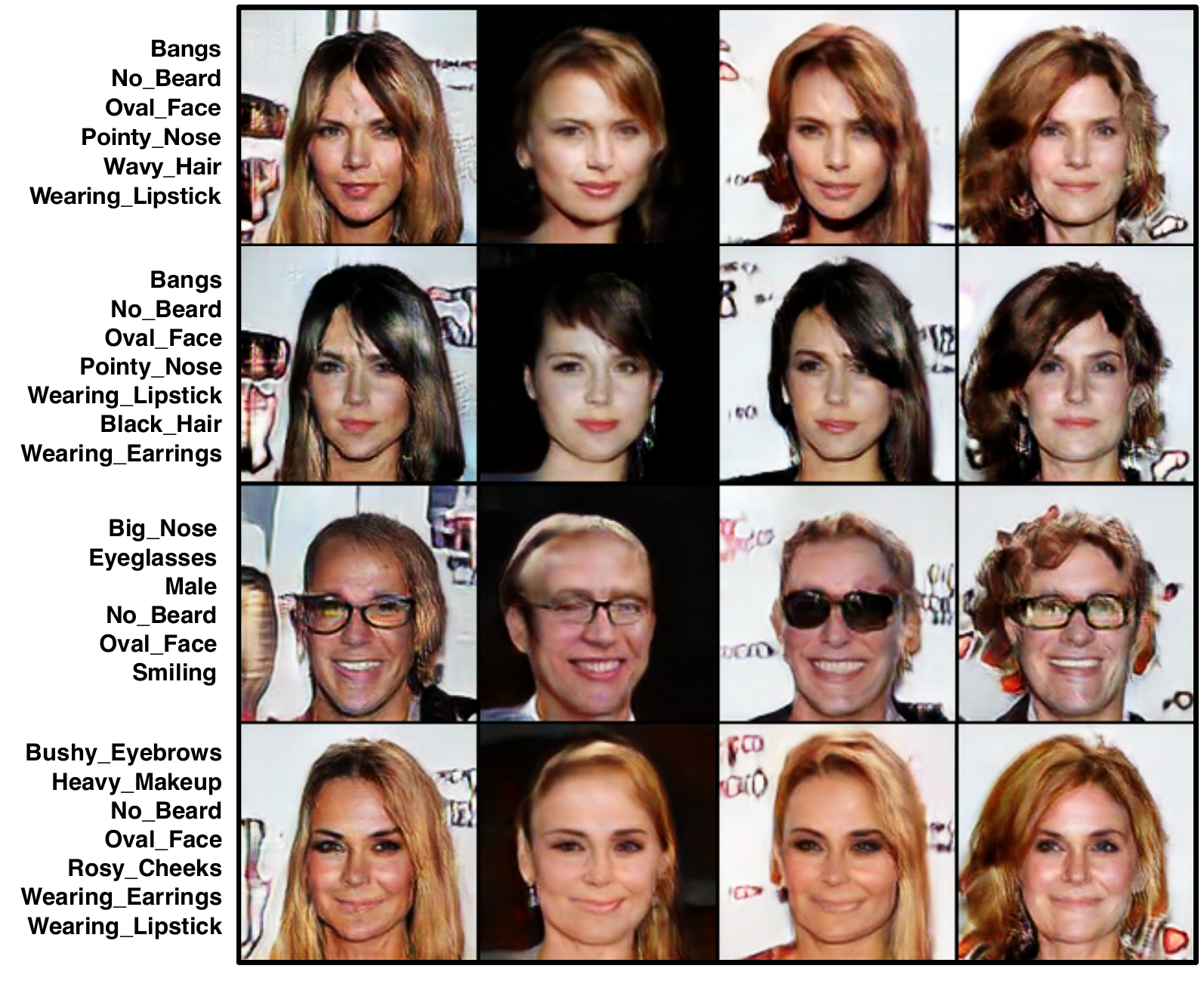}
\caption{\label{fig:attr2faces}Conditional generation given attributes learned by our model in the Attributes-to-Faces task. We sample a set of attributes from the data distribution and generate 4 faces by sampling latent codes from $z_b\sim p(z_b)$.}
\end{figure}

\subsection{Attributes-to-Faces}
In this task, we make use of the CelebA dataset in order map from descriptive facial attributes $A$ to images of faces $B$ and vice-versa. We report both quantitative and qualitative results. For the quantitative results, we follow~\cite{gan2017triangle} and test our models in a semi-supervised attribute prediction setting. We let the model train on all the available data without the pairing information and only train with a small amount of paired data as described in Sec. 3.2. We report Precision (P) and normalized Discounted Cumulative Gain (nDCG) as the two metrics for multi-label classification problems. As an additional baseline, we also train a supervised classifier (which has the same architecture as $\G$) on the paired subset. The results are reported in Table~\ref{tab:attr_recons}. In Fig.~\ref{fig:attr2faces}, we show some generation obtained from the model in the direction attributes to faces. We can see that the model generates reasonable diverse faces for the same set of attributes.

\begin{table}[t!]
\centering
\small
\begin{tabular}{lcc}
    \toprule
    \textbf{Model} & \multicolumn{2}{c}{\textbf{P@10 / NDCG@10}} \\
    		       &   $s =$ 1\% & $s = $ 10\% \\
    \midrule
    Triple-GAN$^\dagger$    & 40.97 / 50.74 & 62.13 / 73.56 \\
    $\Delta$-GAN$^\dagger$  & 53.21 / 58.39 & 63.68 / 75.22 \\
    \midrule
    Baseline Classifier  & 63.36 / 79.25 & 67.34 / 84.21 \\
	AugCGAN              & \textbf{64.38 / 80.59} & \textbf{68.83 / 85.51} \\
    \bottomrule
  \end{tabular}
  \caption{CelebA semi-supervised attribute prediction with supervision $s=$ 1\% and 10\% . $^\dagger$ From \citep{gan2017triangle}.}
  \label{tab:attr_recons}
\end{table}

\section{Conclusion}
% \todo{discuss problems of non-identifiability and disentangling information between noise and input domain.}
In this paper we have introduced the Augmented CycleGAN model for learning many-to-many cross-domain mappings in unsupervised fashion. This model can learn stochastic mappings which leverage auxiliary noise to capture multi-modal conditionals. Our experimental results verify quantitatively and qualitatively the effectiveness of our approach in image translation tasks. Furthermore, we apply our model in a challenging task of learning to map across attributes and faces, and show that it can be used effectively in a semi-supervised learning setting.

% We apply our model on two image translation tasks, and show both quantitatively and qualitatively, that it can learn multi-modal mappings. We further apply our model in a challenging task of learning to map across attributes and faces, and show that it can be used effectively in a semi-supervised learning setting.

\section*{Acknowledgements}
Authors would like to thank Zihang Dai for valuable discussions and feedback. We are also grateful for ICML anonymous reviewers for their comments.

\bibliography{local}
\bibliographystyle{icml2018}

\end{document}